\documentclass[10pt,twocolumn,letterpaper]{article}

\usepackage{cvpr}              

\usepackage[dvipsnames]{xcolor}

\definecolor{cvprblue}{rgb}{0.21,0.49,0.74}
\usepackage[pagebackref,breaklinks,colorlinks,citecolor=cvprblue]{hyperref}
\usepackage{bm}
\usepackage{multirow}
\usepackage{colortbl}
\usepackage{amssymb}
\usepackage{pifont}
\usepackage{arydshln}
\usepackage{enumitem}
\usepackage{lipsum}
\usepackage[T1]{fontenc}

\def\paperID{1361} 
\def\confName{CVPR}
\def\confYear{2025}

\title{CARE Transformer: Mobile-Friendly Linear Visual Transformer via Decoupled Dual Interaction}

\author{
Yuan Zhou\textsuperscript{1}, Qingshan Xu\textsuperscript{1},  Jiequan Cui\textsuperscript{1}, Junbao Zhou\textsuperscript{1}, Jing Zhang\textsuperscript{2}, Richang Hong\textsuperscript{3}, Hanwang Zhang\textsuperscript{1}\\
{\small \textsuperscript{1}Nanyang Technological University \qquad \qquad  \textsuperscript{2}Beihang University \qquad \qquad \textsuperscript{3}Hefei University of Technology}  \\
{\tt\footnotesize \{yuan.zhou, qingshan.xu@ntu.edu.sg, hanwangzhang\}@ntu.edu.sg, zhang\_jing@buaa.edu.cn}
\\ {\tt\footnotesize \{bowmanchow, jiequancui, hongrc.hfut\}@gmail.com}}

\begin{document}
\maketitle

\begin{abstract}
    Recently, large efforts have  been made to design efficient linear-complexity visual Transformers. However, current linear attention models are generally unsuitable to be deployed in resource-constrained mobile devices, due to suffering from either few efficiency gains or significant accuracy drops. In this paper, we propose a new de\textbf{C}oupled du\textbf{A}l-interactive linea\textbf{R} att\textbf{E}ntion (CARE) mechanism, revealing that features' decoupling and interaction can fully unleash the power of linear attention. We first propose an asymmetrical feature decoupling strategy that asymmetrically decouples the learning process for local inductive bias and long-range dependencies, thereby preserving sufficient local and global information while effectively enhancing the efficiency of models. Then, a dynamic memory unit is employed to maintain critical information along the network pipeline. Moreover, we design a dual interaction module to effectively facilitate interaction between local inductive bias and long-range information as well as among features at different layers. By adopting a decoupled learning way and fully exploiting complementarity across features, our method can achieve both high efficiency and accuracy. Extensive experiments on ImageNet-1K, COCO, and ADE20K datasets demonstrate the effectiveness of our approach, e.g., achieving $78.4/82.1\%$ top-1 accuracy on ImagegNet-1K at the cost of only $0.7/1.9$ GMACs. Codes will be released on \href{https://github.com/zhouyuan888888/CARE-Transformer}{https://github.com/zhouyuan888888/CARE-Transformer}.
\end{abstract}

\section{Introduction}
\label{sec:introduction}

Transformers \cite{transformer} were originally designed for natural language processing tasks due to their inherent superiority in global receptive field and parallelizability. In the past few years, they have been successfully introduced to computer vision, including image classification \cite{vit,pyramidvit,swin,cswin}, semantic segmentation \cite{cheng2021per,xie2021segformer,strudel2021segmenter,zhou2024advancing}, object detection \cite{carion2020end,li2022exploring}, and image editing \cite{zhou2025streaming,zhou2025dragnext}. Nevertheless, the quadratic complexity of self-attention severely limits the applicability of Transformers in resource-constrained scenarios, such as on mobile devices.

\begin{figure}[t]
    \centering
    \includegraphics[width=0.9\linewidth]{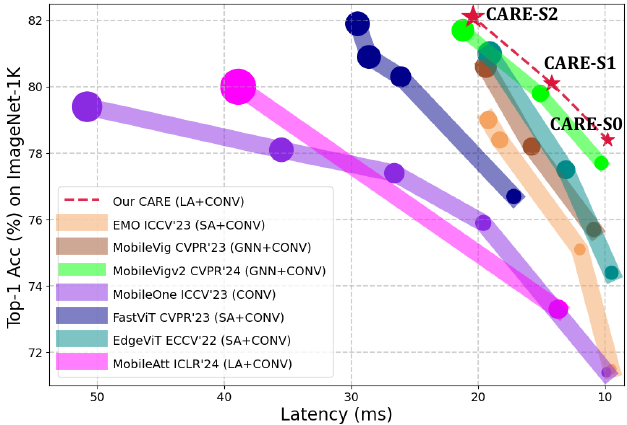}
  
     \caption{Visualized comparison of the balance between accuracy, latency, and GMACs for our CARE Transformers and recent typical mobile-friendly models. In the figure, the larger the marker, the more GMACs the model consumes. ``SA'', ``LA'', and ``GNN'' indicate that the methods are based on Self-Attention, Linear Attention, and Graph Neural Networks. For more details, please refer to Table~\ref{tab:imagenet-1k}.}
     \label{fig:onecol}
\end{figure}

Recently, large efforts have been dedicated to designing efficient linear-complexity visual Transformers, which can be grouped into two  main roadmaps: \emph{i)} limiting the receptive field by using local attention; \emph{ii)} utilizing linear attention to lower down complexity. Local attention improves the efficiency of visual Transformers at the cost of sacrificing their capability to capture long-range dependencies \cite{swin,cswin,neighborhood,pyramidvit,deformable}, which have been proven to play a critical role in learning high-quality feature representations \cite{convnext,replknet}. In contrast to local attention, linear attention can be seen as a more elegant way to solve the quadratic complexity issue \cite{katharopoulos2020transformers,shen2021efficient,choromanski2020rethinking,lu2021soft}. In fact, the quadratic complexity problem in self-attention primarily stems from the use of an explicit similarity measurement between queries and keys. Therefore, linear attention utilizes a kernel trick to remove the softmax function from self-attention and changes the overall computation order, therefore scaling computation complexity to the linear while still ensuring the global receptive field of models. 

\begin{figure}[t]
    \centering
     \includegraphics[width=0.9\linewidth]{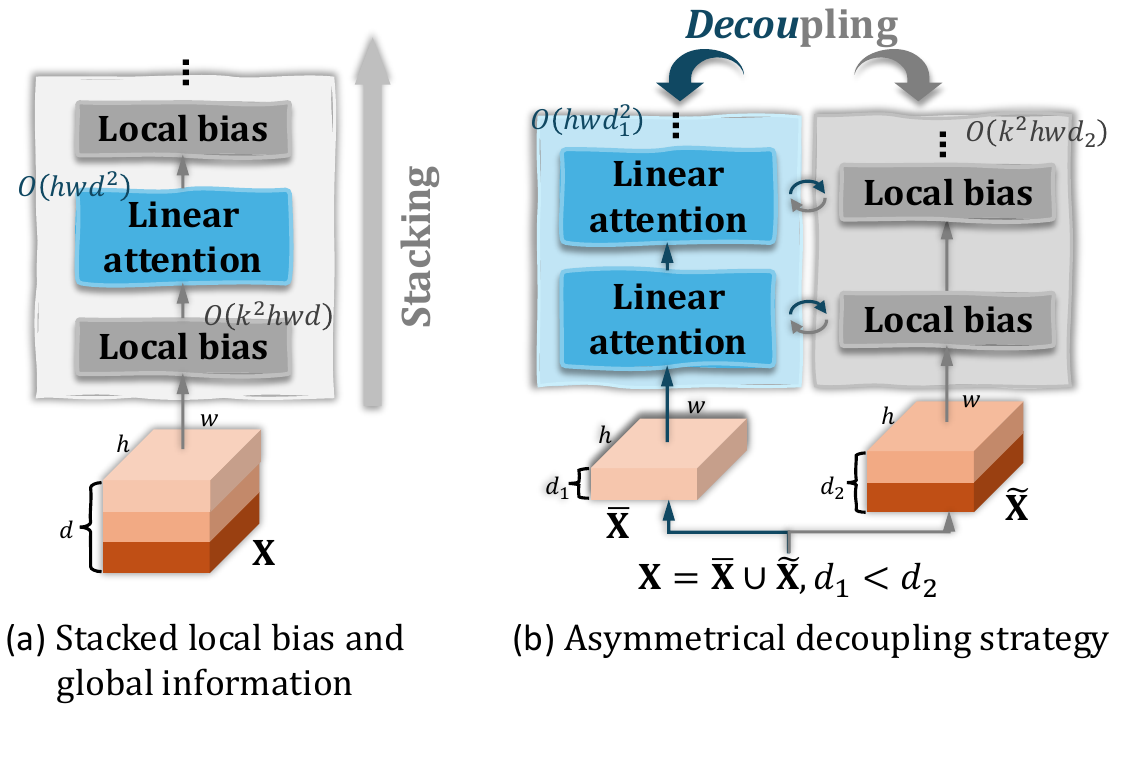}
     \caption{Comparisons between (a) the stacked learning approach \cite{mlla} and (b) our asymmetrical decoupled learning way ($d_1<d_2$) that learns local inductive bias and long-range dependencies separately and further alleviates the quadratic overhead of linear attention to channel dimension, \emph{i.e.,} $\mathrm{O}(hwd_1^2)$. Local inductive bias is learned by using depth-wise convolutions \cite{cpe,cswin,mlla} and long-range dependencies are captured by utilizing linear attention \cite{mlla}.}
     \label{fig:ove}
\end{figure}

A main drawback of linear attention lies in its high entropy property, meaning it struggles to suppress the influence of irrelevant tokens because of not explicitly measuring tokens' relationships \cite{zhang2024hedgehog}. Thus, as shown in Figure \ref{fig:ove} (a), the recent work \cite{mlla} proposed to incorporate linear attention with local inductive bias in a stacked manner, so as to guide models to be aware of global contexts but focus more on important neighboring pixels. The stacked local enhancement can boost the effectiveness of linear attention. But we found that its integration between local inductive bias and long-range information remains inflexible and inefficient, which is realized through the alternate use of convolutions and linear attention, making it still unsuitable to be deployed in resource-constrained scenarios where models' efficiency is critical. This naturally motives us to ask two questions: \emph{\textbf{$\mapsto$Q1.} Is stacking the optimal approach to fully leverage the advantages of local inductive bias and long-range information? \textbf{$\mapsto$Q2.} Is it possible to enhance linear visual Transformers' efficiency and accuracy at the same time?} 

Aiming to answer these two questions, we propose a new \emph{de\textbf{C}oupled du\textbf{A}l-interactive linea\textbf{R} att\textbf{E}ntion (CARE)} mechanism and introduce our CARE Transformers. For the question \emph{$\mapsto$\textbf{Q1}}, we argue that \emph{asymmetrical feature decoupling} can fully unleash the power of linear attention. As illustrated in Figure \ref{fig:ove} (b), by decoupling features in channel dimension, inputs do not need to undergo all the convolution and linear attention operations. Besides, the asymmetry (\emph{i.e.}, $d_1<d_2$) can further alleviate the quadratic overhead of linear attention to channel dimension (\emph{i.e.}, $\mathrm{O}(hwd_1^2)$). Based on the answer to \emph{$\mapsto$\textbf{Q1}}, we argue that by adopting the asymmetrical decoupled learning way and \emph{fully considering complementarity across features}, the question \emph{$\mapsto$\textbf{Q2}} can be solved. To fully exploit the complementarity of features, we first design a dynamic memory unit to preserve the critical information along the network pipeline. Then, we introduce a dual interaction module to effectively facilitate interaction between local bias and long-range dependencies as well as among features at different layers. The asymmetrical decoupling strategy saves computational costs on learning local inductive bias and global information, while cross-feature interaction can flexibly and effectively exploit the information in learned features, thus helping our decoupled learning approach achieve both high efficiency and accuracy. 

All-in-all, the contributions of this paper can be summarized below:
\begin{itemize}
    \item We propose an asymmetrical decoupling strategy that learns local inductive bias and long-range dependencies in a divide-and-conquer way, thereby preserving sufficient local and global cues while further improving models' efficiency.
    \item We design a dynamic memory unit and a dual interaction module to fully leverage the complementarity between local inductive bias and long-range dependencies as well as among features at different layers, helping our decoupled learning approach to achieve both high efficiency and accuracy.
    \item We conduct extensive experiments on ImageNet-1K, ADE20K, and COCO datasets to validate the effectiveness of our approach. The experimental results demonstrate the competitive performance of our models, \emph{e.g.,} achieving $78.4/82.1\%$ top-1 accuracy on ImageNet-1K  with $1.1/2.0ms$ latency on iPhone13 and $0.8/1.5ms$ latency on iPad~Pro.
\end{itemize} 

\section{Related Work}
\label{sec:rel}
We briefly review the research that is relevant to our work in this section. We review local attention in Section \ref{localatt} and introduce the recent advance in linear attention in Section~\ref{linearatt}.

\subsection{Local Attention}
\label{localatt}
Swin \cite{swin} is a typical method based on local attention, where a shifted local attention was introduced to facilitate communication between features across windows. A halo-based local attention was proposed in HaloNet \cite{halonet} to better capture information from the neighborhood of each query. NAT \cite{nat} proposed a neighborhood attention, which can be more dynamic and flexible to aggregate information from neighboring pixels, thereby overcoming the drawback of the non-overlapping window partition. To better capture data-dependent local patterns, a deformable attention and a bi-level routing strategy were introduced by DAT \cite{dat} and BiFormer \cite{biformer}, in order to dynamically search important regions or pixes according to queries. CSWin \cite{cswin} proposed a cross-shaped local attention, which further reduces computational expenses via calculating attention in horizontal and vertical stripes. Using local attention can alleviate the quadratic complexity issue of self-attention but inevitably restricts models' capability to capture long-range information, which is critical for learning high-quality features.

\subsection{Linear Attention}
\label{linearatt}
Nystromformer \cite{nystromformer} proposed to use the matrix approximation strategy Nystrom to approximate self-attention. Performer \cite{choromanski2020rethinking} designed a random orthogonal feature mechanism to provide an unbiased linear-complexity estimation for self-attention. SOFT \cite{lu2021soft} replaced the softmax function with Gaussian kernels, enabling self-attention to be approximated by low-rank matrix decomposition. EfficientAtt \cite{shen2021efficient} proposed to apply softmax to queries and keys separately, therefore decomposing self-attention to the linear. \cite{mobileatt} further reduced the head dimension of linear attention by introducing a head competition mechanism. A focused linear attention was proposed by FLatten \cite{flatten} to further boost the focus ability and diversity of linear attention. SLAB \cite{slab} advanced \cite{flatten} by introducing a re-parameterized BatchNorm, thereby further improving the speed of linear attention models. Agent \cite{agent} combined softmax and linear attention together by introducing agent tokens. Directly approximating self-attention by using linear complexity is difficult. So, recently, MLLA \cite{mlla} enhanced linear attention by leveraging complementarity between local and global information in a stacked manner. Our work takes a step further on MLLA  \cite{mlla} and reveals that features' asymmetrical decoupling and interaction can fully unleash the power of linear attention.

\section{Methodology}
\label{sec:met}

\begin{figure*}[t]
    \centering
     \includegraphics[width=0.71\linewidth]{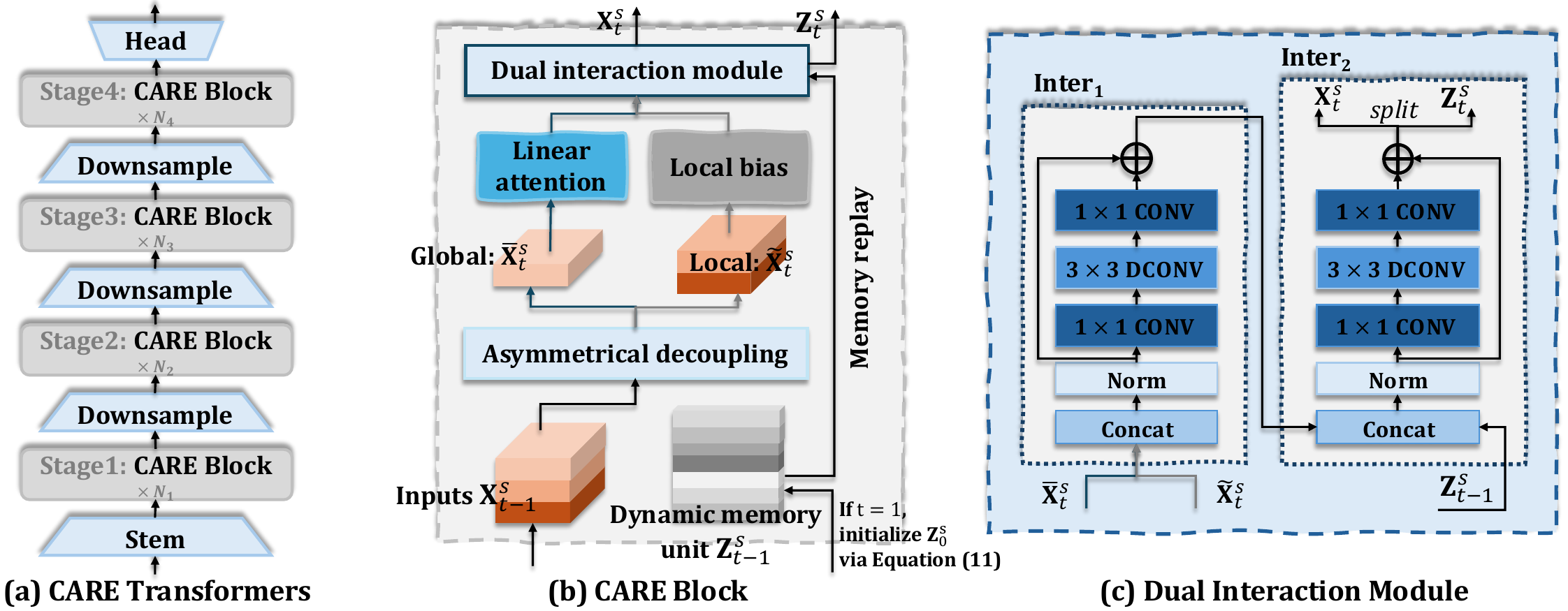}
     \caption{Schematic illustration for CARE Transformers, built by our proposed de\textbf{C}oupled du\textbf{A}l-interactive linea\textbf{R} att\textbf{E}ntion (CARE).}
     \label{fig:sche}
\end{figure*}

\subsection{Preliminary}
\label{sec:pre}
\textbf{Self-Attention}. In a self-attention, the features $\mathbf{X}\in\mathbb{R}^{hw\times d}$ of tokens are first projected into three subspaces, $\mathbf{Q}=\mathbf{X}\mathbf{W}_Q$, $\mathbf{K}=\mathbf{X}\mathbf{W}_K$, and $\mathbf{V}=\mathbf{X}\mathbf{W}_V$, called queries, keys, and values respectively. Then, Equation (\ref{eq:1}) is utilized to calculate the outputs $ \mathbf{O}\in\mathbb{R}^{hw\times d}$ of the self-attention
\begin{align}
    \label{eq:1}
    \mathbf{O}_i = \sum_{j=1}^{N} \frac{\exp \left(\mathbf{Q}_i \mathbf{K}_j^\top \right)}{\sum_{t=1}^{N} \exp \left(\mathbf{Q}_i \mathbf{K}_t^\top \right)} \mathbf{V}_j\mathbf{W}_{O},
\end{align}
where $\mathbf{Q}_i$, $\mathbf{K}_i$, and $\mathbf{V}_i$ represent the query, key, and value about the $i$-th token, and $\mathbf{W}_{O}$ denotes the weights for output projection.

\noindent\textbf{Linear Attention}. In linear attention, a kernel trick $\exp(\mathbf{Q}\mathbf{K}^\top)=\varphi(\mathbf{Q})\varphi(\mathbf{K})^\top$ is further employed to change the computation order of Equation~(\ref{eq:1}) and scale the complexity to the linear,
\begin{align}
    \mathbf{O}_i = \sum_{j=1}^{N} \frac{\varphi(\mathbf{Q}_i)(\varphi(\mathbf{K}_j)^\top \mathbf{V}_j)}{\varphi(\mathbf{Q}_i) \sum_{t=1}^{N}\varphi(\mathbf{K}_t)^\top}\mathbf{W}_{O}.
\end{align}
Recently, \cite{sun2023retentive,yang2023gated} found setting the kernel function $\varphi(\cdot)$ as an identity operator can achieve satisfying performance. Thus, following these methods, we set $\varphi(\cdot)$ as an identity in our work.

\subsection{Motivation}
\label{motivation}
\textbf{Linear Attention with Stacked Local Enhancement}. In order to leverage the advantages of both local and global information, a stacked local enhancement strategy was introduced in \cite{mlla}, which can be described by Equation (\ref{eq:3}):
\begin{align}
    \label{eq:3}
    \mathbf{O} = \mathrm{LocalBias}(\mathrm{LinearAtt}(\mathrm{LocalBias}(\mathbf{X}))),
\end{align}
where $\mathrm{LinearAtt}(\cdot)$ represents a linear attention operation and $\mathrm{LocalBias}(\cdot)$ denotes a local inductive bias learner. Local inductive bias guides models to focus more on neighboring pixels, thus relieving the interference of irrelevant information in learned long-range dependencies. Each local inductive bias learner in \cite{mlla} is implemented by using a depth-wise convolution. We denote the kernel size of these convolutions as $k\times k$, and therefore the computational complexity cost in learning local inductive bias and long-range dependencies is 
\begin{align}
    \Omega=\underbrace{2k^2hwd}_{\text{Local inductive bias}}+\underbrace{\overbrace{4hwd^2}^{\text{Projection}}+\overbrace{2hwd^2}^{\text{\textbf{QKV} Multiplication}}}_{\text{Long-range  dependencies}}. 
\end{align}
\noindent\textbf{$\Game$ Proposition 1.} We argue that the process of the local enhancement can be explicitly divided into two steps: \emph{\textbf{$\mapsto $S1.} learning local inductive bias and long-range information from inputs;} \emph{\textbf{$\mapsto $S2.} fully exploiting the complementarity between them.} 

However, \cite{mlla} does not distinguish these two steps and implicitly unifies them into one pipeline, where the overall process is formulated as the alternate use of convolution and linear attention operations to aggregate local and global information.

\noindent\textbf{$\Game$ Proposition 2.} Unifying these two steps leads to two drawbacks: \emph{\textbf{$\mapsto$D1.} input features should undergo all of the local and global processors, which is the computation bottleneck of the stacked local enhancement;} \emph{\textbf{$\mapsto$D2.} the flexibility of algorithms is damaged severely, as this makes it hard to specifically design a more effective fusion block to facilitate information exchange between local and global features.}

\emph{The above observations motivate us to propose a new divide-and-conquer way to efficiently learn and exploit the information of local inductive bias and long-range dependencies.}

\subsection{Asymmetrical Feature Decoupling: Divide and Conquer!}
\label{sec:free}

Based on the observations in Section \ref{motivation},
we propose an asymmetrical feature decoupling strategy. For the input features $\mathbf{X}\in\mathbb{R}^{hw\times d}$, we first divide them into two parts, represented as $\mathbf{\bar{X}}\in\mathbb{R}^{hw\times d_1}$ and $\mathbf{\widetilde{X}}\in\mathbb{R}^{hw\times d_2}$, where the dimensions satisfy $d_1+d_2=d$. Then, $\mathbf{\bar{X}}$ and $\mathbf{\widetilde{X}}$ are fed to a linear attention operation and a local inductive bias learner, 
\begin{align}
    &\mathbf{\bar{X}}, \mathbf{\widetilde{X}} = \mathrm{Split}(\mathbf{X}, \textit{dim=}1),\\
    &\mathbf{\bar{X}} = \mathrm{LinearAtt}(\mathbf{\bar{X}}),\\
    &\mathbf{\widetilde{X}} = \mathrm{LocalBias}(\mathbf{\widetilde{X}}),
\end{align}
where $\mathbf{\widetilde{X}}$ and $\mathbf{\bar{X}}$ indicate the learned features about local inductive bias and long-range dependencies. Thereby, local and global information can be learned in a divide-and-conquer way. The computational complexity cost in learning local inductive bias and long-range dependencies can be reduced to 
\begin{align}
\label{eq:8}
\Omega=2\lambda_2k^2hwd+4\lambda_1hwd^2+2\lambda_1hwd^2,
\end{align}
where $\lambda_2=\frac{d_2}{d}$ and $\lambda_1=(\frac{d_1}{d})^2$ are scaling factors about the computational complexity after using our decoupled learning way.

\noindent\textbf{$\Game$ Proposition 3.} \emph{Linear attention has quadratic complexity to channel dimension, \emph{i.e.}, $\mathrm{O}(hwd^2)$. Features should be decoupled in an asymmetrical way, meaning $d_1+d_2=d$ and $d_1<d_2$, which further boosts the efficiency of models.}

\noindent\textbf{$\Game$ Proof 1.} \emph{The asymmetrical setting $d_1<d_2$ can reduce computation complexity further. Let $d_2-d_1=\Delta$. Therefore, we have $d_1=\frac{d-\Delta}{2}$ and $d_2=\frac{d+\Delta}{2}$ due to $d_1+d_2=d$ and $d_2-d_1=~\Delta$. We can rewrite Equation (\ref{eq:8}) as follows:}
\begin{align}
    \Omega(\Delta)&=k^2hw(d+\Delta)+\frac{3}{2}hw(d-\Delta)^2.
\label{eq:9}
\end{align}
\emph{Accordingly,}
\begin{align}
    \Omega(\Delta_1)-\Omega(\Delta_2)=\frac{3}{2}hw(\Delta_1-\Delta_2)(\frac{2}{3}k^2+\Delta_1+\Delta_2-2d)
\end{align}
\emph{Let $\Delta_1>0$ and $\Delta_2=0$, we have $\Omega(\Delta_1)-\Omega(0)<0$ as $\Delta_1<d$ and the kernel size generally obeys $k^2<<d$, proving our asymmetrical setting ($\Delta_1>0$) has less complexity compared to the symmetrical scenario ($\Delta_2=0$).}

\subsection{Decoupled Dual-Interactive Linear Attention}
Based on the asymmetrical feature decoupling strategy, we propose a new \emph{de\textbf{C}oupled du\textbf{A}l-interactive linea\textbf{R} att\textbf{E}ntion (CARE)}, which is shown in Figure~\ref{fig:sche}~(b). Our CARE fully considers the complementarity of features, and helps the decoupled learning approach to achieve both high accuracy and efficiency.

\noindent\textbf{Notions}. For better explanations, in this part, we use $\mathbf{X}^{s}_{t}\in\mathbb{R}^{hw\times d}$ and $\mathbf{Z}^{s}_{t}\in\mathbb{R}^{hw\times d'}$ to represent the features and the dynamic memory unit processed by the $t$-th CARE block of the $s$-th stage. 

\noindent\textbf{Dynamical Memory Unit}. Features from different layers have their own advantages and are complementary to each other \cite{chen2018progressively,piao2019depth}. Therefore, we introduce a dynamic memory unit in our CARE to dynamically record learned features along the network. For the first CARE block of each stage $s$, the dynamic memory unit $\mathbf{Z}^{s}_{0}$ is built by both considering the features and the stored memory from the previous stage~$(s-1)$
\begin{align}
    &\mathbf{Z}^{s}_{0} = \mathrm{CONV}_{2\times 2}(\mathbf{X}^{s-1}_{-1}\oplus \mathbf{Z}^{s-1}_{-1}, \textit{stride=}2),
\end{align}
where $\mathbf{X}^{s-1}_{-1}$ and $\mathbf{Z}^{s-1}_{-1}$ indicate the features and the memory unit updated in the last block of the stage $(s-1)$, and $\oplus$ denotes a concatenation operation. We set the stride of the $2\times 2$ convolution $\mathrm{CONV}_{2\times 2}(\cdot)$ as $2$ to keep the consistency between the resolutions of the features in the stage $(s-1)$ and~$s$. In this way, the features from previous layers can be replayed and interacted with the features in the subsequent stages.

\noindent\textbf{Dual Interaction}. CARE considers the dual interaction between features, \emph{i.e., i) the interaction between local and global information,} and \emph{ii) the interaction between features of different layers}, which can be described by Equation (\ref{eq:12}):
\begin{align}
    \label{eq:12}
    &\mathbf{X}^{s}_t, \mathbf{Z}^s_t = \mathrm{Inter}_2(\mathrm{Inter}_1(\mathbf{\bar{X}}^s_{t},\mathbf{\widetilde{X}}^s_{t}),\mathbf{Z}^s_{t-1}),
\end{align}
 where $\mathbf{\bar{X}}^s_{t}\in\mathbb{R}^{hw\times d_1}$ and $\mathbf{\widetilde{X}}^s_{t}\in\mathbb{R}^{hw\times d_2}$ are global and local features learned in the $t$-th CARE block of the $s$-th stage as described in Equation (\textcolor{red}{5}), (\textcolor{red}{6}), and (\textcolor{red}{7}), and $\mathbf{X}^{s}_t\in\mathbb{R}^{hw\times d}$ denotes the features that contain both local and global information after the dual interaction. The first interaction function $\mathrm{Inter}_1(\cdot)$ facilitates the information exchange between local and global features, while the second interaction function $\mathrm{Inter}_2(\cdot)$ further considers the interaction between features in different layers and dynamically updates the memory information $\mathbf{Z}^s_t$. By conducting the dual interaction, the information in different types of features learned by the network can be fully considered, which is beneficial for improving accuracy \cite{zhou2023few,fei2025path}.

\noindent\textbf{Dual Interaction Module}. As shown in Figure~\ref{fig:sche}~(c), a new dual interaction module is designed to facilitate interaction between features, where the two interaction blocks are utilized to implement the two interaction functions $\mathrm{Inter}_1(\cdot)$ and $\mathrm{Inter}_2(\cdot)$ respectively. Each interaction block can be described by
\begin{align}
    \label{eq:13}
    &\mathrm{Inter_*}(\mathbf{x},\mathbf{y})=\mathrm{CONV}_{1\times1,3\times3,1\times1}(\mathrm{Norm}(\mathbf{x}\oplus \mathbf{y})),
\end{align}
where $*=~1$~or~$2$ represents the index of the interaction block and $\mathrm{CONV}_{1\times1,3\times3,1\times1}(\cdot)$ denotes a sequence of convolutions. Specifically, we first concatenate $\mathbf{x}$ and $\mathbf{y}$ together and utilize a normalization layer $\mathrm{Norm}(\cdot)$ to normalize the inputs. Then, a $1\times 1$ convolution is used to perform feature interaction in channel domain and map the inputs to a latent high dimension space with the expansion set as $4$. Also, a $3\times 3$ depth-wise convolution is adopted to conduct feature interaction in spatial domain. Finally, a $1\times 1$ convolution is further utilized to enable feature interaction across channels and map the latent features to the original space. 

\subsection{CARE Transformers}
Based on our CARE, we further introduce CARE Transformers. As illustrated in Figure \ref{fig:sche} (a), our CARE Transformers totally contain four stages, each of which is made up of several CARE blocks. We implement the stem layer by using a $4\times 4$ convolution with the stride set as~$4$. For CARE-S0, CARE-S1, and CARE-S2, the number of attention blocks in each stage are set as $\langle 2, 4, 8, 4 \rangle$, $\langle 3, 6, 10, 6\rangle$, and $\langle 3, 6, 10, 6\rangle $. Accordingly, the dimensions of features are set as $\langle 24, 48, 96, 192\rangle $, $\langle 24, 48, 96, 192\rangle $, and $\langle 24, 48, 144, 288\rangle $. For all of our models, we adopt the asymmetrical setting $d_1=\frac{1}{2}d_2$, and set the dimension of the dynamic memory unit as $d'=d_1$. To help models capture local inductive bias in multiscale receptive filed, we implement the local bias learner $\mathrm{LocalBias}(\cdot)$ using multiple convolutions, \emph{i.e.}, a $3\times 3$ and a $7\times 7$ depth-wise convolutions, utilized in the decoupling manner as well. That means just like Equation (\textcolor{red}{5}), (\textcolor{red}{6}), and (\textcolor{red}{7}), we split local features into two parts, utilize $3\times 3$ and $7\times 7$ convolutions to process them respectively, and then concatenate them together to form the final outputs of $\mathrm{LocalBias(\cdot)}$. Also, following \cite{fastvit,mobileatt}, linear attention operations are not applied to the first two stages, and a $1\times 11$ and a $11\times 1$ depth-wise convolution are employed to aggregate long-range information.

\begin{table*}[t]
    \caption{The performance of our models on ImageNet-1K. The latency on iPhone13 (A15) and iPad Pro (M2) are measure by using xcode and CoreMLTools, where CPU, GPU, and NPU are all considered. The batch size is set as 1 for the latency tested on the iPhone13, iPad Pro, and Intel i9-10940X CPU device, and 64 for the RTX 4090 GPU card. We test our models on the standard $224\times 224$ input resolution. ``SA'', ``LA'', and ``GNN'' indicate that the methods are based on Self-Attention, Linear Attention, and Graph Neural Networks respectively.}
    \label{tab:imagenet-1k}
    \centering
    \vskip 0.05in    
    \begin{minipage}[t]{1\linewidth}
        \resizebox{\linewidth}{!}{
        \setlength{\tabcolsep}{4mm}{
        \renewcommand\arraystretch{1.205}
        \begin{tabular}{c l c c c c c c c c c c}
            \toprule
             &  &  &  &  & & \multicolumn{4}{c}{\textbf{Latency ($ms$) $\downarrow$}} & \\
            \cline{7-10}
            & \textbf{Method}  & \textbf{Ref.} & \textbf{Type} & \textbf{GMACs} $\downarrow$  & \textbf{Params ($M$)} $\downarrow$ & \textbf{iPhone13} & \textbf{iPad Pro} & \textbf{Intel i9} & \textbf{RTX 4090} & \textbf{Top-1 Acc ($\%$)} $\uparrow$ \\
            \midrule
            \midrule
            \multirow{11}{*}{\rotatebox{90}{$\mathit{\bm{0<\textbf{\textit{GMACs}}<0.8}}$}} & MobileNetV2-$1.0$ \cite{mobilenet_v2} & CVPR' 18 & CONV & 0.3 & 3.5  & 1.0 & 0.7 & 1.1 & 9.3 & 71.8\\
            & MobileOne-S0 \cite{mobileone} & CVPR' 23 & CONV & $0.3$ & $2.1$  & $1.1$ & $0.8$ & $0.7$ & $9.9$ &  $71.4$ \\
            & EMO-1M \cite{emo} & ICCV' 23 & SA+CONV & $0.3$ & $1.3$ & $1.6$  & $1.2$ & $7.4$ & $9.5$ & $71.5$ \\
            & EMO-2M \cite{emo} & ICCV' 23 & SA+CONV & $0.4$ & $2.3$ & $2.0$ & $1.5$ & $11.3$ & $12.0$ &  $75.1$ \\
            & MobileNetV2-$1.4$ \cite{mobilenet_v2} & CVPR' 18 & CONV  & $0.6$ & $6.1$ & $1.4$ & $8.3$ & $1.8$ & $13.4$ & $74.7$ \\ 
            & EdgeViT-XXS \cite{edgevit} & ECCV' 22 & SA+CONV & $0.6$ & $4.1$ & $23.6$ & $17.1$ & $4.5$ & $9.5$ & $74.4$ \\
            & PVT-v2-B0 \cite{pyramidvit_v2}  & CVM' 22 & SA & $0.6$ & $3.7$ & $2.5$ & $1.7$ & $14.3$ & $13.5$ &  $70.5$ \\
            & MobileAtt-PVT-v2-B0 \cite{mobileatt} & ICLR' 24 & LA & $0.6$ & $3.5$ & $3.2$ & $2.4$ & $20.8$ & $17.7$ & $71.5$ \\
            & MobileViGv2-T \cite{mobileViGv2} & CVPR' 24 & GNN+CONV & $0.6$ & $5.6$ & $1.1$ & $0.7$ & $3.8$ & $10.3$ &  $77.7$ \\
            & FastViT-T8 \cite{fastvit} & CVPR' 23 & SA+CONV  & $0.7$ & $3.6$ & $1.3$ & $0.9$ & $11.1$ & $17.2$ &  $76.7$ \\ 
            & MobileViG-T \cite{mobileVIG} & CVPR' 23 & GNN+CONV & $0.7$ & $5.2$ & $1.0$ & $0.7$ & $3.8$ & $10.9$ &  $75.7$ \\
            \hdashline
            \rowcolor{gray!20} & Our CARE-S0 & - & LA+CONV & $0.7$ & $7.3$ & $1.1$ & $0.8$ & $4.3$ & $9.8$ &  $78.4$ \\
            \midrule
            \multirow{12}{*}{\rotatebox{90}{$\mathit{\bm{0.8\leq\textbf{\textit{GMACs}}<1.5}}$}} & MobileOne-S1 \cite{mobileone} & CVPR' 23 & CONV  & $0.8$ & $4.8$ & $1.3$ &  $0.9$ & $1.5$ & $19.6$ &  $75.9$ \\
            & EMO-5M \cite{emo} & ICCV' 23 & SA+CONV & $0.9$ & $5.1$ & $2.7$ & $1.9$ & $17.6$ & $18.3$ &  $78.4$ \\
            & MobileViGv2-S \cite{mobileViGv2}  & CVPR' 24 & GNN+CONV  & $0.9$ & $7.7$ & $1.3$ & $0.9$ & $5.6$ & $15.1$ & $79.8$ \\
            & EMO-6M \cite{emo} & ICCV' 23 & SA+CONV  & $1.0$ & $6.1$ & $2.9$ & $2.1$ & $21.2$ & $19.2$ &  $79.0$ \\
            & MobileViG-S \cite{mobileVIG} & CVPR' 23 & GNN+CONV & $1.0$ & $7.2$ & $1.2$ & $0.8$ & $5.6$ & $15.8$ &  $78.2$ \\
            & EdgeViT-XS \cite{edgevit} & ECCV' 22 & SA+CONV & $1.1$ & $6.7$ & $34.0$ & $20.2$ & $6.6$ & $13.1$ & $77.5$ \\
            & Agent-DeiT-T \cite{agent} & ECCV' 24 & SA & $1.2$ & $6.0$ & $3.1$ & $2.9$ & $24.9$ & $12.5$ &  $74.9$ \\
            & MobileAtt-DeiT-T \cite{mobileatt} & ICLR' 24 & LA & $1.2$ & $5.7$ & $3.1$ & $2.7$ & $8.1$ & $13.7$ & $73.3$ \\
            & DeiT-T \cite{deit} & ICML' 21 & LA & $1.3$ & $5.7$ & $2.1$ & $1.4$ & $3.9$ & $10.5$ &  $72.2$ \\
            & MobileOne-S2 \cite{mobileone} & CVPR' 23 & CONV & $1.3$ & $7.8$ & $1.4$ & $1.1$ & $2.2$ & $26.6$ &  $77.4$ \\
            & SLAB-DeiT-T \cite{slab} & ICLR' 24 & LA & $1.3$ & $6.2$ & $1.6$ & $1.2$ & $6.2$ & $10.4$ &  $73.6$ \\
            & FastViT-T12 \cite{fastvit} & CVPR' 23 & SA+CONV & $1.4$ & $6.8$ & $2.1$ & $1.3$ & $16.1$ & $26.1$ & $80.3$ \\
            \hdashline
            \rowcolor{gray!20} & Our CARE-S1 & - & LA+CONV & $1.0$ & $9.6$ & $1.4$ & $1.1$ & $6.6$ & $14.2$ &  $80.1$ \\
   
            \midrule
            \multirow{22}{*}{\rotatebox{90}{$\mathit{\bm{1.5\leq\textbf{\textit{GMACs}}}}$}} & MobileViG-M \cite{mobileVIG} & CVPR' 23 & GNN+CONV & $1.5$ & $14.0$ & $1.6$ & $1.1$ & $6.7$ & $19.4$ & $80.6$ \\
            & MobileViGv2-M \cite{mobileViGv2} & CVPR' 24 & GNN+CONV & $1.6$ & $15.4$ & $1.6$ & $1.2$ & $7.2$ & $21.2$ & $81.7$ \\
            & FastViT-S12 \cite{fastvit} & CVPR' 23 & SA+CONV & $1.8$ & $8.8$ & $2.3$ & $1.4$ & $17.8$ & $28.6$ &  $80.9$ \\
            & EdgeViT-S \cite{edgevit} & ECCV' 22 & SA+CONV & $1.9$ & $11.1$ & $53.8$ & $32.7$ & $10.1$ & $19.1$ &  $81.0$ \\
            & MobileOne-S3 \cite{mobileone} & CVPR' 23 & CONV & $1.9$ & $10.1$ & $1.7$ & $1.3$ & $3.2$ & $35.5$ &  $78.1$ \\
            & FastViT-SA12 \cite{fastvit} & CVPR' 23 & SA+CONV & $1.9$ & $10.9$ & $2.4$ & $1.5$ & $18.5$ & $29.5$ &  $81.9$  \\
            & SLAB-PVT-T \cite{slab} & ICLR' 24 & LA & $1.9$ & $13.4$ & $2.5$ & $1.6$ & $19.8$ & $26.9$ & $76.0$ \\
            & FLatten-PVT-T \cite{flatten} & CVPR' 23 & LA & $2.0$ & $12.2$ & $3.4$ & $2.3$ & $20.2$ & $29.4$ &  $77.8$ \\
            & Agent-PVT-T \cite{agent} & ECCV' 24 & SA & $2.0$ & $11.6$ & $3.2$ & $2.1$ & $59.7$ & $23.4$ & $78.4$ \\
            & FLatten-PVTv2-B1 \cite{flatten} & CVPR' 23 & LA & $2.2$ & $19.2$ & $4.1$ & $3.3$ & $41.9$ & $35.9$ & $79.5$ \\
            & MobileOne-S4 \cite{mobileone} & CVPR' 23 & CONV & $3.0$ & $14.8$ & $2.4$ & $1.7$ & $7.2$ & $50.8$ & $79.4$ \\
            & MobileAtt-PVT-v2-B2 \cite{mobileatt} & ICLR' 24 & LA & $3.8$ & $21.1$ & $9.4$ & $7.6$ & $80.2$ & $73.5$ & $82.6$ \\
            & PVT-v2-B2 \cite{pyramidvit_v2} & CVM' 22 & SA & $4.0$ & $25.4$ & $5.6$ & $4.4$ & $55.4$ & $44.6$ & $82.1$ \\ 
            & DeiT-S \cite{deit} & ICML' 21 & SA & $4.2$ & $22.0$ & $3.8$ & $2.4$ & $11.2$ & $25.1$ & $79.8$ \\   
            & MobileAtt-DeiT-S \cite{mobileatt} & ICLR' 24 & LA & $4.2$ & $22.0$ & $4.8$ & $3.7$ & $19.8$ & $38.9$ & $80.0$ \\
            & MLLA-T \cite{mlla} & NeurIPS' 24 & LA+CONV & $4.2$ & $25.0$ & $5.1$ & $3.8$ & $21.3$ & $51.5$ &  $83.5$ \\
            & Agent-Swin-T \cite{agent} & ECCV' 24 & SA & $4.5$ & $29.0$ & $11.7$ & $8.8$ & $25.3$ & $43.6$ & $82.6$ \\
            & Swin-T \cite{swin} & ICCV' 21 & SA & $4.5$ & $29.0$ & $12.6$ & $9.4$ & $26.3$ & $47.8$ & $81.3$ \\
            & ConvNeXt-T \cite{convnext} & CVPR' 22 & CONV & $4.5$ & $29.0$ & $90.9$ & $68.9$ & $15.1$ & $38.2$ & $82.1$ \\
            & FLatten-Swin-T \cite{flatten} & CVPR' 23 & LA & $4.5$ & $29.0$ & $12.8$ & $9.2$ & $24.8$ & $46.3$ &  $82.1$ \\
            & SLAB-Swin-T \cite{slab} & ICLR' 24 & LA & $4.5$ & $29.0$ & $12.6$ & $8.3$ & $23.8$ & $43.2$ & $81.8$ \\

            \hdashline
            \rowcolor{gray!20} & Our CARE-S2 & - & LA+CONV & $1.9$ & $19.5$ & $2.0$ & $1.5$ & $9.4$ & $20.4$ & $82.1$ \\
            \bottomrule

         \end{tabular}}}
    \end{minipage}
    \vspace{-0.4cm}
\end{table*}

\begin{table*}[t]
    \caption{The performance of our models on Object Detection (OB), Instance Segmentation (IS), and Semantic Segmentation (SS) tasks. Following \cite{mobileatt,mobileVIG,mobileViGv2}, the latency and GMACs of backbones are included in the table, measured on the $512\times 512$ input resolution. In addition, for ensuring fairness, all models are compared on the head Mask R-CNN on OB and IS, and the head Semantic FPN on SS.}
    \label{tab:coco}
    \centering
    \vskip 0.05in    
    \begin{minipage}[t]{1\linewidth}
        \resizebox{\linewidth}{!}{
        \setlength{\tabcolsep}{2.5mm}{
        \renewcommand\arraystretch{1.205}
        \begin{tabular}{lccccc|ccc|ccc|cc}
            \toprule
            & & & & & & \multicolumn{3}{c|}{\textbf{OB} $\uparrow$} & \multicolumn{3}{c|}{\textbf{IS} $\uparrow$} & \textbf{SS} $\uparrow$ \\
            \cline{7-14}
            \textbf{Backbone} & \textbf{Ref.} & \textbf{Type} & \textbf{GMACs} $\downarrow$  & \textbf{iPhone13 ($ms$)} $\downarrow$  & \textbf{RTX 4090 ($s$)} $\downarrow$  & $\bm{\mathrm{AP}}^{b}$ & $\bm{\mathrm{AP}}^{b}_{50}$ & $\bm{\mathrm{AP}}^{b}_{75}$ & $\bm{\mathrm{AP}}^{m}$ & $\bm{\mathrm{AP}}^{m}_{50}$ & $\bm{\mathrm{AP}}^{m}_{75}$ & \textbf{mIoU} \\
            \midrule
            \midrule
            EfficientFormer-L1 \cite{efficientformer} & NeurIPS' 22 & CONV & $6.9$ & $4.8$ & $0.09$ &  $37.9$ & $60.3$ & $41.0$ & $35.4$ & $57.3$ & $37.3$ & $38.9$ \\
            MobileViG-M \cite{mobileVIG} & CVPR' 23 & GNN+CONV & $7.8$ & $5.0$ & $0.12$ & $41.3$ & $62.8$ & $45.1$ & $38.1$ & $60.1$ & $40.8$ & n/a \\
            MobileViGv2-M \cite{mobileViGv2} & CVPR' 24 & GNN+CONV & $8.5$ & $5.5$ & $0.14$ & $42.5$ & $63.9$ & $46.3$ & $38.8$ & $60.8$ & $41.7$ & $42.9$ \\
            ResNet18 \cite{resnet} & CVPR' 16 & CONV & $9.5$ & $3.2$ & $0.04$ &  $34.0$ & $54.0$ & $36.7$ & $31.2$ & $51.0$ & $32.7$ & $32.9$ \\
            FastViT-SA12 \cite{fastvit} & CVPR' 23 & SA+CONV & $8.0$ & $6.9$ & $0.14$ &  $38.9$ & $60.5$ & $42.2$ & $35.9$ & $57.6$ & $38.1$ & $38.0$ \\
            PoolFormer-S12 \cite{metaformer} & CVPR' 22 & CONV & $9.5$ & $5.7$ & $0.09$ & $37.3$ & $59.0$ & $40.1$ & $34.6$ & $55.8$ & $36.9$ & $37.2$ \\
            SLAB-PVT-T \cite{slab} & ICLR' 24 & LA & $10.4$ & $14.2$ & $0.13$ & $36.5$ & $59.0$ & $39.2$ & $34.4$ & $55.7$ & $36.5$ & n/a \\
            FLatten-PVT-T  \cite{flatten} & CVPR' 23 & LA & $10.4$ & $18.6$ & $0.14$ &  $38.2$ & $61.6$ & $41.9$ & $37.0$ & $57.6$ & $39.0$ & $37.2$ \\
            Agent-PVT-T \cite{agent} & ECCV' 24 & SA & $10.7$ & $17.6$ & $0.12$ &  $41.4$ & $64.1$ & $45.2$ & $38.7$ & $61.3$ & $41.6$ & $40.2$ \\
            PoolFormer-S24 \cite{metaformer} & CVPR' 22 & CONV & $17.7$ & $9.8$ & $0.17$ & $40.1$ & $62.2$ & $43.4$ & $37.0$ & $59.1$ & $39.6$ & $40.3$ \\

            EfficientFormer-L3 \cite{efficientformer} & NeurIPS' 22 & CONV & $21.0$ & $10.3$ & $0.20$ & $41.4$ & $63.9$ & $44.7$ & $38.1$ & $61.0$ & $40.4$ & $43.5$ \\
            ResNet50 \cite{resnet} & CVPR' 16 & CONV & $21.4$ & $7.6$ & $0.14$ & $38.0$ & $58.6$ & $41.4$ & $34.4$ & $55.1$ & $36.7$ & $36.7$ \\
            MLLA-T \cite{mlla} & NeurIPS' 24 & LA+CONV & $22.0$ & $25.5$ & $0.29$ &  $46.8$ & $69.5$ & $51.5$ & $42.1$ & $66.4$ & $45.0$ & n/a \\
            Swin-T \cite{swin} & CVPR' 22 & SA & $24.2$ & $28.2$ & $0.28$ & $42.2$ & $64.4$ & $46.2$ & $39.1$ & $64.6$ & $42.0$ & $41.5$ \\
            FLatten-Swin-T \cite{flatten} & CVPR' 23 & LA & $24.2$ & $28.6$ & $0.28$ & $44.2$ & $67.3$ & $48.5$ & $40.2$ & $63.8$ & $43.0$ & $44.8$ \\
            Agent-Swin-T \cite{agent} & ECCV' 24 & SA & $24.2$ & $26.8$ & $0.26$ & $44.6$ & $67.5$ & $48.7$ & $40.7$ & $64.4$ & $43.4$ & $46.7$ \\
            SLAB-Swin-T \cite{slab} & ICLR' 24 & LA & $24.2$ & $26.4$ & $0.26$ &  $43.9$ & $66.5$ & $48.1$ & $40.1$ & $63.3$ & $43.1$ & n/a \\
            ResNet101 \cite{resnet} & CVPR' 16 & CONV & $40.8$ & $11.5$ & $0.22$ &  $40.4$ & $61.1$ & $44.2$ & $36.4$ & $57.7$ & $38.8$ & $38.8$ \\
            \hdashline
            \rowcolor{gray!20} Our CARE-S0 & - & LA+CONV & $3.8$ & $4.1$ & $0.07$ &  $40.3$ & $62.2$ & $43.8$ & $36.8$ & $58.8$ & $39.2$ & $38.5$ \\
            \rowcolor{gray!20} Our CARE-S1 & - & LA+CONV & $5.4$ & $5.5$ & $0.11$ & $41.5$ & $63.4$ & $44.8$ & $37.8$ & $60.2$ & $39.9$ & $41.0$ \\
            \rowcolor{gray!20} Our CARE-S2 & - & LA+CONV & $10.1$ & $7.3$ & $0.14$ &  $43.1$ & $64.4$ & $47.0$ & $39.0$ & $61.6$ & $41.9$ & $43.5$ \\
            \bottomrule
        \end{tabular}}}
    \end{minipage}
    \vspace{-0.4cm}
\end{table*}

\section{Experiments}
\label{sec:exp}

We conduct extensive experiments to demonstrate the effectiveness of our method. We first compare our models with counterparts on the classification task of ImageNet-1K \cite{imagenet}, and then introduce the results of our models on downstream COCO \cite{coco} and ADE20K \cite{ade20k} datasets. Finally, we provide ablation studies to analyze the impact of each component of our models.

\subsection{Image Classification}
\textbf{Implementation Details}. ImageNet-1K \cite{imagenet} is one of the most famous image classification benchmarks, which includes about $1.3$ M training images and $50$ K test images. On ImageNet-1K, we build our codes by using PyTorch \cite{pytorch} and timm \cite{timm}, and use AdamW \cite{loshchilov2017decoupled} to update model parameters where the initial learning is set as $2e^{-3}$. We train our models on $12$ RTX 4090 GPU cards for $300$ epochs with the batch size set as $128$ per GPU. Data augmentation strategies, including standard random resized crop, horizontal flip, RandAugment \cite{cubuk2020randaugment}, Mixup \cite{zhang2017mixup}, CutMix \cite{yun2019cutmix}, random erasing \cite{zhong2020random}, and color jitter are used during training. Also, label smoothing \cite{szegedy2016rethinking}, stochastic depth \cite{huang2016deep}, and weight decay are employed to regularize our training process.

\noindent\textbf{Experimental Results}. The results on the ImageNet-1K dataset are summarized in Table \ref{tab:imagenet-1k}. According to these results, we can draw the following conclusions. Firstly, at the same level of computational costs, our models clearly have higher accuracy. For example, at the cost of $0.7$ GMACs, our CARE-S0 can achieve $78.4\%$ accuracy, higher than that of MobileViG-T and FastViT-T8 by $2.7\%$ and $1.7\%$. At the cost of $1.9$ GMACs, the accuracy of our CARE-S2 is $1.1\%$ and $4.0\%$ higher than that of EdgeViT-S and MobileOne-S3. Secondly, compared with the models having the same level of accuracy, our CARE Transformers have obviously higher efficiency. For example, CARE-S2 just costs half the GMACs of MLLA-T and has $2.5\times$ faster speed on iPhone13 and iPad Pro, yet its accuracy is only lower than that of MLLA-T by $1.4\%$. These experimental results indicate that our models can achieve a better balance between accuracy and efficiency, and demonstrate their superiority in being deployed in resource-constrained mobile devices.

\subsection{Object Detection \& Instance Segmentation}

\textbf{Implementation Details}. COCO \cite{coco} is a frequently-used dataset in the object detection and instance segmentation task, which contains around $118$ K training samples and $5$ K validation images. On this benchmark, we build our models by using PyTorch \cite{pytorch} and mmdetection \cite{mmdet}. Moreover, we utilize the AdamW \cite{loshchilov2017decoupled} optimizer to train our models for $12$ epochs where the initial learning rate is set as $2e^{-4}$ and the batch size is set as 24. We use the models pre-trained on the ImageNet-1K dataset to initialize the parameters of our backbones. Following the previous works \cite{mobileVIG,mobileViGv2}, we implement the detection head by using Mask R-CNN \cite{he2017mask}.

\noindent\textbf{Experimental Results}. The results in Table \ref{tab:coco} demonstrate the effectiveness of our method again. On the object detection task, our CARE-S0 outperforms FLatten-PVT-T, SLAB-PVT-T, FastViT-SA12, PoolFormer-S12, and PoolFormer-S24 in both accuracy and efficiency. Although CARE-S1 has obviously less GMCAs, it can still achieve comparable accuracy to MobileViG-M, MobileViGv2-M, Agent-PVT-T, and Swin-T. Our CARE-S2 just costs half of the GMACs of MLLA-T and has $3\times$ faster speed on iPhone13 and $2\times$ faster speed on the RTX 4090 GPU card, but the $\mathrm{AP}^b$ of MLLA-T is only $3.7\%$ higher that of our model. Similar results can be found on the instance segmentation task as well. For instance, our CARE-S2 has obviously higher efficiency than Swin-T, Flatten-Swin-T, Agent-Swin-T, and SLAB-Swin-T, but can still maintain comparable accuracy. These results demonstrate the superiority of our method in resource-constrained scenarios.
 
\subsection{Semantic Segmentation}
\textbf{Implementation Details}. We also evaluate the performance of our approach on the semantic segmentation task of the ADE20K dataset, which includes about $20$ K training samples and $2$ K test images from 150 categories. On this dataset, we build our models by using PyTorch \cite{pytorch} and mmsegmentation \cite{mmseg}. We use the checkpoints pre-trained on the ImageNet-1K dataset to initialize our backbone networks, which are then incorporated with the segmentation head Semantic FPN \cite{kirillov2019panoptic}. We use the AdamW \cite{loshchilov2017decoupled} optimizer to train our models where the initial learning rate is set as $2e^{-4}$ with a polynomial decay schedule and the batch size is set as 48.

\noindent\textbf{Experimental Results}. The results in Table \ref{tab:coco} indicate that our approach can still achieve competitive performance on semantic segmentation. For example, despite that our CARE-S1 just costs half of the GMACs of Agent-PVT-T, FLatten-PVT-T, and PoolFormer-S12, its mIoU is still higher than that of these counterparts by $0.83\%$, $3.8\%$, and $3.8\%$. The accuracy of our CARE-S2 is comparable to that of the models Swin-T, FLatten-Swin-T, and Agent-Swin-T. But CARE-S2 has obviously higher efficiency. For example, the latency of our CARE-S2 is less than that of these three counterparts by $20.9ms$, $21.3ms$, and $19.5ms$ on iPhone13, and $0.14s$, $0.14s$, and $0.12s$ on the RTX 4090 GPU card.

\begin{table}[t]
    \caption{Ablation studies for our asymmetrical feature decoupling strategy, conducted on CARE-S2 on ImageNet-1K. ``\textit{w/}~Asym'' indicates using the full asymmetrical feature decoupling. ``\textit{w/}~Sym'' represents decoupling features in a symmetrical way. ``\textit{w/} Sta'' represents that our asymmetrical decoupling strategy is not employed, and local and global information are learned in the stacked manner. ``\textit{w/o} Local'' indicates the baseline where linear attention is not incorporated with the guidance of local inductive bias.}
    \label{tab:decoupling}
    \centering
    \vskip 0.05in    
    \begin{minipage}[t]{1\linewidth}
        \resizebox{\linewidth}{!}{
        \setlength{\tabcolsep}{0.8mm}{
        \renewcommand\arraystretch{1.205}
        \begin{tabular}{c c c c c c c c c c c}
            \toprule
              &   &  &  \multicolumn{2}{c}{\textbf{Latency ($ms$)} $\downarrow$} &  \\
             \cline{4-5}
            \textbf{Experiment} &  \textbf{GMACs} $\downarrow$ & \textbf{Params ($M$)} $\downarrow$ & \textbf{iPhone13}  & \textbf{RTX 4090} & \textbf{Top-1 Acc ($\%$)} $\uparrow$ \\
            \midrule
            \midrule
            \textit{w/} Sym & $2.0$ & $19.7$ & $2.2$ & $21.8$ & $82.1$ \\ 
            \textit{w/} Sta & $2.1$ & $20.9$ & $2.6$ & $24.8$ & $81.4$ \\
            \textit{w/o} Local & $2.0$ & $20.7$ & $2.5$ & $24.0$ & $77.3$ \\
            \hdashline
            \rowcolor{gray!20}\textit{w/} Asym & $1.9$ & $19.5$ & $2.0$ & $20.4$ & $82.1$  \\
            \bottomrule
         \end{tabular}}}
    \end{minipage}
    \vspace{-0.4cm}
\end{table}

\subsection{Ablation Studies}
\label{sec:abl}
\textbf{Ablation on the Asymmetrical Feature Decoupling}. As can be seen from Table \ref{tab:decoupling}, the full asymmetrical decoupling strategy (``\textit{w/} Asym'') can help the model to achieve better performance in both efficiency and accuracy. In contrast, the use of the stacked learning approach (``\textit{w/} Sta'') lowers the accuracy and efficiency of the model at the same time, \emph{e.g.,} only achieving $81.4\%$ top-1 accuracy on ImageNet-1K and need to cost $2.6ms$ latency on iPhone13 and $24.8ms$ latency on the RTX 4090 GPU card. That indicates the asymmetrical decoupling and dual interaction of our CARE can effectively and efficiently exploit the information in learned features, which is better than the alternate use of convolution and linear attention operations. Also, our asymmetrical decoupled learning strategy is still better than the symmetrical one (``\textit{w/} Sym''), \emph{e.g.}, these two strategies achieve comparable accuracy but our asymmetrical decoupled learning approach has lower costs in terms of all GAMCs, parameters, and latency. We also compare our method with the baseline (``\textit{w/o} Local'') where linear attention is not incorporated with local inductive bias. Without the guidance of local inductive bias, the accuracy of the baseline is obviously lower than that of ours. Meanwhile, its efficiency is still lower than that of our model, as our asymmetrical feature decoupling can further reduce the quadratic computational overhead of linear attention to channels, demonstrating the effectiveness of our method in solving \emph{\textbf{$\mapsto$Q2}}, \emph{i.e.,} improving the accuracy and efficiency of linear visual Transformers  at the same time. 

\begin{table}[t]
    \caption{Ablation studies for the dual interaction module, conducted on CARE-S0 on ImageNet-1K. ``Inter$_1$'' and ``Inter$_2$'' denote the first and the second interaction block. ``Mem'' indicates whether the memory information is considered in ``Inter$_2$''.}
    \label{tab:ablation}
    \centering
    \vskip 0.05in    
    \begin{minipage}[t]{1\linewidth}
        \resizebox{\linewidth}{!}{
        \setlength{\tabcolsep}{1mm}{
        \renewcommand\arraystretch{1.205}
        \begin{tabular}{c c c c c c c c c c c}
            \toprule
            \multicolumn{3}{c}{\textbf{Experiment}} & & & \multicolumn{2}{c}{\textbf{Latency ($\bm{ms}$)} $\downarrow$}  & \\
            \cline{1-3} \cline{6-7}
            \textbf{Inter$_1$} & \textbf{Inter$_2$} & \textbf{Mem} &  \textbf{GMACs} $\downarrow$ & \textbf{Params ($M$)} $\downarrow$ & \textbf{iPhone13} & \textbf{RTX 4090} & \textbf{Top-1 Acc ($\%$)} $\uparrow$ \\
            \midrule
            \midrule
            $\times$ & \checkmark & \checkmark & 0.5 & 5.5 & 1.0 & 7.4 & 76.2 \\
            \checkmark & $\times$ & \checkmark & 0.3 & 2.7 & 0.9 & 7.0 & 70.9 \\
            \checkmark & \checkmark & $\times$ & 0.5 & 5.4 & 1.1 & 8.5 & 76.5 \\
            \hdashline
            \rowcolor{gray!20} \checkmark & \checkmark & \checkmark & 0.7 & 7.3 & 1.1 & 9.8 & 78.4\\
            \bottomrule
        \end{tabular}}}
    \end{minipage}
    \vspace{-0.4cm}
\end{table}

\vspace{0.2cm}

\noindent\textbf{Ablation on the Dual Interaction Module.} We provide the ablation study for our dual interactive module in Table~\ref{tab:ablation}. According to these results, we have the following observations. The use of the full dual interactive module can achieve the highest top-1 accuracy. While removing the first interaction block (\emph{i.e.,} Inter$_1$) can slightly improve the efficiency of the model, this leads to an obvious $2.2\%$ accuracy drop. Discarding the second interaction block (\emph{i.e.,} Inter$_2$) further decreases the accuracy to $70.9\%$, demonstrating the importance of fully exploiting the complementarity between local inductive bias, long-range dependencies, and dynamically stored memory. In Table \ref{tab:ablation}, we also study the influence of the dynamical memory unit. We find removing the dynamical memory unit from Inter$_2$ causes a $1.9\%$ accuracy drop, indicating that exploiting the information of different layers guides models to achieve better performance.

\vspace{0.2cm}

\noindent\textbf{Ablation on CARE Transformers}. We provide the ablation study for our CARE Transformers in Table \ref{tab:care}. We find that replacing $1\times11$ and $11\times1$ convolutions with linear attention in the first two layers causes a slight accuracy drop, as there exists more noisy or redundant information in the shallow layers of neural networks but linear attention's high entropy property makes it difficult to suppress the influence of noisy or redundant information. The results in the table also demonstrate that learning local inductive bias in multiscale receptive field can help the model achieve better performance without sacrificing models' efficiency obviously.

\begin{table}[t]
    \caption{Ablation studies for our CARE Transformers. ``Early CONV'' indicates that long-range information in the first two stages are captured by a $1\times11$ and a $11\times1$ convolution. `` MS'' indicates learning local inductive bias in multiscale receptive field.}
    \label{tab:care}
    \centering
    \vskip 0.05in    
    \begin{minipage}[t]{1\linewidth}
        \resizebox{\linewidth}{!}{
        \setlength{\tabcolsep}{2mm}{
        \renewcommand\arraystretch{1.205}
        \begin{tabular}{c c c c c c c c c c c}
            \toprule
            \multicolumn{3}{c}{\textbf{Experiment}}  &   &  &   &  \\
            \cline{1-3}
            \textbf{Backbone} & \textbf{Early CONV} & \textbf{MS} & \textbf{GMACs} $\downarrow$ & \textbf{Params ($M$)} $\downarrow$   & \textbf{Top-1 Acc ($\%$)} $\uparrow$ \\
            \midrule
            \midrule

            CARE-S0 & \checkmark & \checkmark & $0.72$ & $7.34$ & 78.4 \\ 
            CARE-S0 & $\times$ & \checkmark & $0.72$ & $7.34$ & 78.3 \\ 
            CARE-S0 & \checkmark & $\times$ & $0.71$ & $7.32$ & 78.1 \\ 
            \hdashline
            CARE-S1 & \checkmark & \checkmark & $1.03$ & $9.60$ & 80.1 \\ 
            CARE-S1 & $\times$ & \checkmark & $1.03$ & $9.60$ & 79.9 \\ 
            CARE-S1 & \checkmark & $\times$ & $1.02$ & $9.57$ & 79.8 \\ 
            \hdashline
            CARE-S2 & \checkmark & \checkmark & $1.93$ & $19.48$ & 82.1 \\ 
            CARE-S2 & $\times$  & \checkmark & $1.93$ & $19.48$ & 82.0 \\ 
            CARE-S2 & \checkmark & $\times$ & $1.92$ & $19.43$ & 81.8 \\ 

            \bottomrule
        \end{tabular}}}
    \end{minipage}
\end{table}

\section{Conclusion}
\label{sec:clu}
In this paper, we propose a new \emph{de\textbf{C}oupled du\textbf{A}l-interactive linea\textbf{R} att\textbf{E}ntion (CARE)} mechanism, which reveals that features' decoupling and interaction can fully unleash the power of linear attention. We propose to symmetrically decouple the learning process for local inductive bias and long-range dependencies, thereby preserving sufficient local and global information while further enhancing the efficiency of models. Moreover, we design a dynamic memory unit and a dual interaction module aiming to fully exploit the complementarity between local inductive bias and long-range dependencies as well as among features at different layers. By adopting a decoupled learning way and fully exploiting complementarity across features, our method can achieve both high efficiency and accuracy, showing its superiority in being deployed in resource-constrained mobile devices.

\noindent\textbf{Limitation}. There are two limitations in our current work. Firstly, we did not utilize NAS to search for the optimal architecture of our CARE Transformers. Secondly, we did not apply our method to models with relatively larger size due to our limited GPU resources; however, we believe that the proposed CARE mechanism can also perform well with large models. We will study these two points in the future.

\clearpage
{
\small
\bibliographystyle{ieeenat_fullname}
\bibliography{main.bib}
}


\end{document}